\def\year{2020}\relax
\documentclass[letterpaper]{article} 
\usepackage{aaai20}  
\usepackage{times}  
\usepackage{helvet} 
\usepackage{courier}  
\usepackage[hyphens]{url}  
\usepackage{graphicx} 
\usepackage{graphicx}  
\usepackage{amsmath,amssymb} 
\usepackage{subfigure} 
\usepackage{flafter} 
\title{Reinforcement Learning with Convolutional Reservoir Computing}
\author{Han-ten Chang, Katsuya Futagami\\ 
Graduate school of Systems and Information Engineering University of Tsukuba\\
1-1-1 Tennodai, Tsukuba, Ibaraki 305-8573, Japan\\
\{s1820554, s1820559\}@s.tsukuba.ac.jp 
}

\begin{document}

\maketitle

\begin{abstract}
Recently, reinforcement learning models have achieved great success, mastering complex tasks such as Go and other games with higher scores than human players. Many of these models store considerable data on the tasks and achieve high performance by extracting visual and time-series features using convolutional neural networks (CNNs) and recurrent neural networks, respectively. However, these networks have very high computational costs because they need to be trained by repeatedly using the stored data. In this study, we propose a novel practical approach called reinforcement learning with convolutional reservoir computing (RCRC) model. The RCRC model uses a fixed random-weight CNN and a reservoir computing model to extract visual and time-series features. Using these extracted features, it decides actions with an evolution strategy method. Thereby, the RCRC model has several desirable features: (1) there is no need to train the feature extractor, (2) there is no need to store training data, (3) it can take a wide range of actions, and (4) there is only a single task-dependent weight parameter to be trained. Furthermore, we show the RCRC model can solve multiple reinforcement learning tasks with a completely identical feature extractor.
\end{abstract}

\section{Introduction}
\noindent Recently, reinforcement learning (RL) models have achieved great success, mastering complex tasks such as Go \cite{silver2016mastering} and other games \cite{DBLP:journals/corr/MnihKSGAWR13,DBLP:journals/corr/abs-1803-00933,kapturowski2018recurrent} with higher scores than human players. Many of these models use convolutional neural networks (CNNs) to extract visual features directly from the environment state images \cite{DBLP:journals/corr/abs-1708-05866}. Some models use recurrent neural networks (RNNs) to extract time-series features and achieved higher scores \cite{hausknecht2015deep}. 

However, these deep neural networks (DNNs) based models are often very computationally expensive in that they train networks weights by repeatedly using a large volume of past playing data and task-rewards. Certain techniques can alleviate these costs, such as the distributed approach \cite{mnih2016asynchronous,DBLP:journals/corr/abs-1803-00933} which efficiently uses multiple agents, and the prioritized experienced replay \cite{schaul2015prioritized} which selects samples that facilitate training. However, the cost of a series of computations, from data collection to action determination, remains high.

The world model \cite{NIPS2018_7512} can also reduce computational costs by completely separating the training processes between the feature extraction model and the action decision model.  The world model trains the feature extraction model in a rewards-independent manner by using variational auto-encoder (VAE) \cite{2013arXiv1312.6114K,2014arXiv1401.4082J} and mixture density network combined with an RNN (MDN-RNN)  \cite{DBLP:journals/corr/Graves13}. After extracting the environment state features, it uses an evolution strategy method called the covariance matrix adaptation evolution strategy (CMA-ES) \cite{doi:10.1162/106365601750190398,DBLP:journals/corr/Hansen16a} to train the action decision model. The world model can achieve outstanding scores in famous RL tasks. The separation of these two models results in the stabilization of feature extraction and reduction of parameters to be trained based on task-rewards. 

From the success of the world model, it is implied that in the RL feature extraction process, it is important to extract the features that express the environment state sufficiently rather than features trained to get higher rewards. Adopting this idea, we propose a new method called ``reinforcement learning with convolutional reservoir computing (RCRC)". The RCRC model is inspired by the reservoir computing. 

Reservoir computing \cite{lukovsevivcius2009reservoir} is a kind of RNNs, and the model weights are set to random. One of the reservoir computing models, the echo state network (ESN)  \cite{jaeger2001echo,jaeger2004harnessing} is used to solve time-series tasks such as future value prediction. For this, the ESN extracts features for the input signal based on the dot product of the input signal and fixed random-weight matrices generated without training. Surprisingly, features obtained in this manner are expressive enough to understand the input, and complex tasks such as chaotic time-series prediction can be solved by using them as the input for a linear model. In addition, the ESN has solved various tasks in multiple fields such as time-series classification  \cite{tanisaro2016time,ma2016functional} and Q-learning-based RL \cite{szita2006reinforcement}. Similarly, in image classification, the model that uses features extracted by the CNN with fixed random-weights as the ESN input achieves high accuracy classification with a smaller number of parameters \cite{Tong2018}.

Based on the success of the fixed random-weight models, the RCRC model extracts the visual features of the environment state using fixed random-weight CNN, and, using these features as the ESN input, extracts time-series features of the environment state transitions. After extracting the environment state features, we use CMA-ES \cite{doi:10.1162/106365601750190398,DBLP:journals/corr/Hansen16a} to train a linear transformation from the extracted features to the actions, as in the world model. This model architecture results in the omission of the training process of feature extractor and reduced computational costs; there is also no need to store past playing data. Furthermore, we show that the RCRC model can solved multiple RL tasks with the completely identical structure and weights feature extractor. 

Our contributions in this study are as follows:
\begin{itemize}
       \item We developed a novel and widely applicable approach to extract visual and time-series features of an RL environment state using fixed random-weights networks feature extractor with no training. 
	\item We developed the RCRC model that doesn't need to store data and to train feature extractor.
	\item We showed that the RCRC model can solve different tasks with training only a single task-dependent weight matrix and using the completely identical feature extractor.	
\end{itemize}

\section{Related Work}
\subsection{Reservoir Computing}
\begin{figure}[t]
  \centering
  \includegraphics[width=0.9\columnwidth]{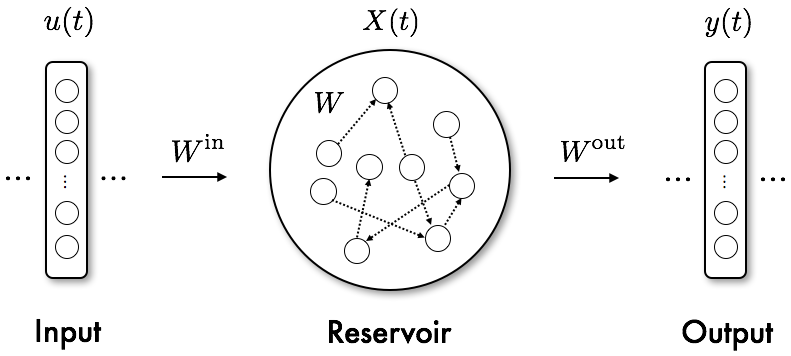}
  \caption{Reservoir Computing overview for the time-series prediction task.}
  \label{fig:fig1}  
\end{figure}
Reservoir computing is one of the RNNs and it extracts features of the input without training for the feature extraction process. In this study, we focus on a reservoir computing model, ESN \cite{jaeger2001echo,jaeger2004harnessing}. The ESN was initially proposed to solve time-series tasks \cite{jaeger2001echo} and is regarded as an RNN model \cite{lukovsevivcius2009reservoir,lukovsevivcius2012practical}. 

Let the $N$-length, $D_u$-dimensional input signal be $u = \{u(1), u(2), ..., u(t),...,u(N)\} \in \mathbb{R}^{N \times D_u}$ and the signal that added one bias term to input signal be $U = [u;1] = \{U(1),U(2),...,U(T),...,U(N)\} \in \mathbb{R}^{N \times (D_u+1)}$. Let [;] be a vector concatenation. The ESN gets features called the reservoir state $X = \{X(1), ..., X(t),...,X(N)\} \in \mathbb{R}^{N \times D_x}$ as follows: 
\begin{eqnarray}
	\tilde{X}(t+1) & = & f(W^{\text{in}}U(t) + WX(t)))\\
	X(t+1) & = & (1-\alpha)X(t) + \alpha\tilde{X}(t+1)
\end{eqnarray}
where the matrices $W^{\text{in}} \in \mathbb{R}^{(D_u+1) \times D_x}$ and $W \in \mathbb{R}^{D_x \times D_x}$ are sampled from a probability distribution such as a Gaussian distribution, and $f$ is the activation function which is applied element-wise. As the activation function, $linear$ and $tanh$ functions are generally used; it is also known that changing the activation function according to the task improves accuracy \cite{inubushi2017reservoir,DBLP:journals/corr/abs-1905-09419}. The leakage rate $\alpha \in [0,1]$ is a hyperparameter that tunes the weight between the current and the previous values, and $W$ has two major hyperparameters called sparsity and spectral radius. The sparsity is the ratio of 0 elements in matrix $W$ and the spectral radius is a memory capacity parameter which is calculated by the maximal absolute eigenvalue of $W$. 

Finally, the ESN estimates the target signal $y = \{y(1), y(2), ..., y(t),...,y(N)\} \in \mathbb{R}^{N \times D_y}$ as 
\begin{eqnarray}
y(t) = W^{\text{out}} [X(t); U(t); 1].
\end{eqnarray}
The weight matrix $W^{\text{out}} \in \mathbb{R}^{D_y \times (D_x + D_u + 1)}$ is estimated by a linear model such as ridge regression. An overview of reservoir computing is shown in Figure\ref{fig:fig1}.

The unique feature of the ESN is that the two matrices $W ^{\text{in}}$ and $W$ are randomly generated from a probability distribution and fixed. Therefore, the training process in the ESN consists only of a linear model to estimate $W ^{\text{out}}$, hence the ESN has very low computational cost. In addition, the reservoir state reflects complex dynamics despite being obtained by random matrix transformation, and it is possible to use it to predict complex time-series by a simple linear transformation \cite{jaeger2001echo,verstraeten2007experimental,DBLP:journals/corr/GoudarziBLTS14}. Because of the low computational cost and high expressiveness of the extracted features, the ESN is also used to solve other tasks such as time-series classification \cite{tanisaro2016time,ma2016functional}, Q-learning-based RL \cite{szita2006reinforcement} and image classification \cite{Tong2018}.

\subsection{World Models}
The world model \cite{NIPS2018_7512} is one of the RL models that separates the training of the feature extraction model and the action decision model to train the model more efficiently. It uses VAE \cite{2013arXiv1312.6114K,2014arXiv1401.4082J} and MDN-RNN \cite{DBLP:journals/corr/Graves13} as feature extractors. They are trained in a reward-independent manner with randomly played 10000 episodes data. As a result, in the feature extraction process, the reward-based parameters are omitted, and there remains a single weight parameter to be trained that decides the action. The weight is trained by CMA-ES \cite{doi:10.1162/106365601750190398,DBLP:journals/corr/Hansen16a}. Although the feature extraction model is trained in a reward-dependent manner, the world model achieves outstanding scores in an RL task {\tt CarRacing-v0} \cite{carracingv0}. Furthermore, the world model can be trained to predict the next environment state, thus it can generate the environment by itself. By training the action decision model in this self-play environment, the world model solved an RL task {\tt DoomTakeCover-v0} \cite{Kempka2016ViZDoom,DoomTakeCover}.

CMA-ES is one of the evolution strategy methods used to optimize some parameters using a multi-candidate search generated from a multivariate normal distribution $\mathcal{N}(m, \sigma^2C)$. The parameters $m$, $\sigma$, and $C$ are updated with a formula called the evolution path. The evolution paths are updated according to the previous evolution paths and the evaluation scores of each candidate. As CMA-ES updates parameters using only the evaluation scores calculated by actual playing, it can be used regardless of whether the actions of the environment are continuous or discrete values \cite{doi:10.1162/106365601750190398,DBLP:journals/corr/Hansen16a}. Furthermore, the training can be faster because it can be parallelized by the number of solution candidates.

The world model reduces the computational cost and accelerates the training process by separating training processes of models and applying CMA-ES. However, in the world model, it is necessary to independently optimize VAE, MDN-RNN, and CMA-ES each other. Furthermore, in optimizing the feature extractor models, they need to save considerable data and be trained by repeatedly using those data.

\begin{figure*}[t]
  \includegraphics[keepaspectratio, width=0.85\textwidth]{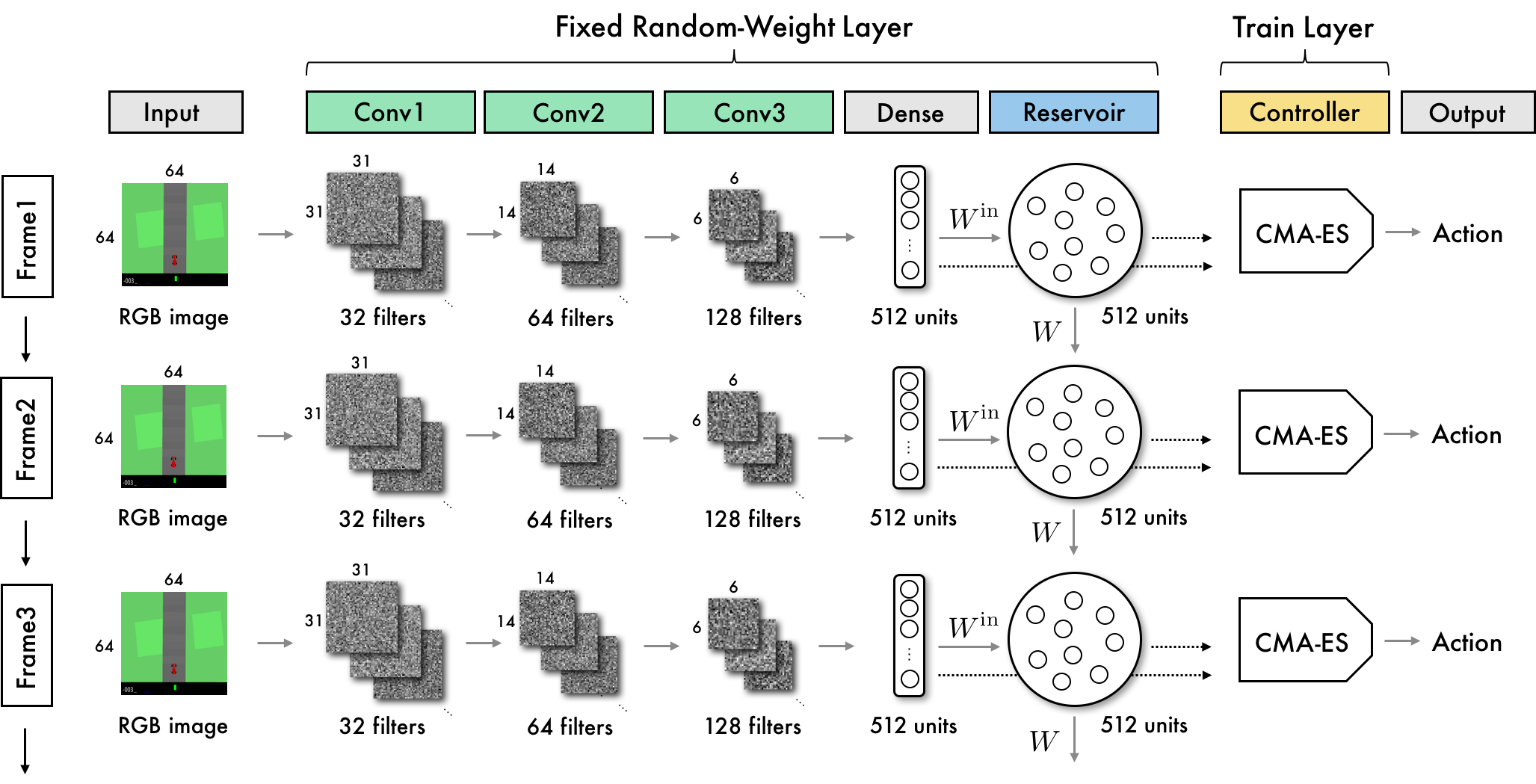}
  \centering
  \caption{RCRC overview to decide the action for {\tt CarRacing-v0}: the feature extraction layer are called the convolutional reservoir computing layer, and the weights are sampled from Gaussian distributions and then fixed. For {\tt DoomTakeCover-v0} we use the completely identical convolutional reservoir computing layer in {\tt CarRacing-v0}. In both tasks, we train only a single weight in the controller layer to take task-dependent actions. }
  \label{fig:fig2} 
\end{figure*}

\section{Proposal Model}
\subsection{Basic Concept}
The world model \cite{NIPS2018_7512} extracts visual features and time-series features of environment states by using VAE \cite{2013arXiv1312.6114K,2014arXiv1401.4082J} and MDN-RNN \cite{DBLP:journals/corr/Graves13} without using task-rewards. The model achieves outstanding scores through the linear transformation of these features. This implies that in RL tasks, it requires features that sufficiently express the environment state, rather than features trained to get higher rewards.

We thus focus on extracting features that sufficiently express environment states by fixed random-weights networks. Using fixed random-weights networks as feature extractor has some advantages, such as no need for both training feature extractor and storing data, while being able to sufficiently extract features. For example, a simple CNN with fixed random-weights can extract visual features and achieve high accuracy in image classification \cite{Tong2018}. Although the MDN-RNN weights are fixed in the world model, it can achieve high performance \cite{Corentin2018rnnfix}. In the ESN, the model can predict complex time-series using features extracted by random matrices transformations \cite{jaeger2001echo,verstraeten2007experimental,DBLP:journals/corr/GoudarziBLTS14}. Therefore, it can be considered that CNNs can extract visual features and ESN can extract time-series features, even if their weights are random and fixed. From this hypothesis, we propose the RCRC model, which includes both fixed random-weight CNN and ESN.
 
\subsection{Proposal Model Overview}
The RCRC model is composed of three layers: the untrained CNN layer, the reservoir computing layer and the controller layer. In the first layer, it extracts visual features by using a fixed random-weight CNN. In the second layer, it uses transitions of the visual features extracted in the first layer as input to the ESN to extract the time-series features. In the two layers above that collectively called the convolutional reservoir computing layer, visual and time-series features are extracted with no training. In the final layer, a single weight matrix of a linear transformation from the outputs of the convolutional reservoir computing layer to the actions is trained. A model overview is shown in Figure~\ref{fig:fig2}.

In the previous study, there is a similar world model-based approach \cite{DBLP:journals/corr/abs-1906-08857} that uses fixed random-weights in VAE and a LSTM \cite{hochreiter1997long}. However, this approach is ineffective in solving {\tt CarRacing-v0} \cite{carracingv0}. In the training process, the best average score over 20 randomly created tracks of each generation was less than 200. However, as mentioned further on, we achieved an average score above 900 over 100 randomly created tracks in {\tt CarRacing-v0} by taking the reservoir computing knowledge in the RCRC model. We also solved {\tt DoomTakeCover-v0} \cite{Kempka2016ViZDoom,DoomTakeCover} by using the convolutional reservoir computing layer whose structure and weights are completely identical in {\tt CarRacing-v0}.

The characteristics of the RCRC model are as follows:
\begin{itemize}
\item The computational cost of the RCRC model is very low because visual and time-series features of environment states are extracted using a convolutional reservoir computing layer whose weights are fixed and random.
\item In the RCRC model, only a single weight matrix in the controller layer needs to be trained because the feature extraction model (the convolutional reservoir computing layer) and the action training model (the controller layer) are separated.
\item The RCRC model can take a wide range of actions regardless of continuous or discrete, because the model training process is based on the scores measured by actual playing.
\item Past data storage is not required, as neither the convolutional reservoir computing layer nor the controller layer needs to be trained by the past data as in back-propagation.
\item The convolutional reservoir computing layer can be applied to other tasks without further training for feature extractor because the layer's weights are fixed with task-independent random weights.
\end{itemize}

\subsection{Convolutional Reservoir Computing layer}
In the convolutional reservoir computing layer, the visual and time-series features of the environment state images are extracted by a fixed random-weight CNN and an ESN-based method which has fixed random-weights, respectively. A study using each image's features that are extracted by the fixed random-weight CNN as input to the ESN has been previously conducted, and has shown its ability to classify MNIST dataset \cite{mnistdata} with high accuracy \cite{Tong2018}. Based on this, we developed a novel and practical approach to solve various RL tasks. By taking advantage of the RL characteristic that the current environment state and the action determine the next environment state, the RCRC model updates the reservoir state with current and previous environment state features. This updating process enables the reservoir state to have time-series features.

More precisely, consider the $D_\text{conv}$-dimensional visual features extracted by the fixed random-weight CNN for $t$-th environment state image $X_{\text{conv}}(t) \in \mathbb{R}^{D_\text{conv}}$ and the $D_\text{esn}$-dimensional reservoir state $X_{\text{esn}}(t) \in \mathbb{R}^{D_\text{esn}}$. The reservoir state $X_{\text{esn}}$ is time-series features and updated as follows:
\begin{eqnarray}
\tilde{X}_{\text{esn}}(t+1) & = & f(W^{\text{in}}X_{\text{conv}}(t) + WX_{\text{esn}}(t)))\\
X_{\text{esn}}(t+1) & =&  (1-\alpha)X_{\text{esn}}(t) + \alpha\tilde{X}_{\text{esn}}(t+1).
\end{eqnarray}
This updating process has no training necessity, and is very fast, because $W^{\text{in}}$ and $W$ are random matrices sampled from the probability distribution and fixed. 

\subsection{Controller layer}
The controller layer decides actions by using the output of the convolutional reservoir computing layer, $X_{\text{conv}}$ and $X_{\text{esn}}$. Let $t$-th environment state input vector which added one bias term be $S(t) = [X_{\text{conv}}(t); X_{\text{esn}}(t); 1]\in \mathbb{R}^{D_{\text{conv}}+D_{\text{esn}}+1}$. We suppose that the feature $S(t)$ has sufficient expressive information about the environment states and it can take action by a linear combination of $S(t)$. Therefore, we obtain action $A(t) \in \mathbb{R}^{N_{\text{act}}}$ as follows:
\begin{eqnarray}
  \tilde{A}(t) & = & W^\text{out}S(t)\\
  A(t) & = & g(\tilde{A}(t)) 
\end{eqnarray}
where $W^{\text{out}} \in \mathbb{R}^{(D_{\text{conv}}+D_{\text{esn}}+1) \times N_{\text{act}}}$ is the weight matrix and the scalar $N_{\text{act}}$ is the number of actions in the task; $g$ is the function which adjusts $\tilde{A}(t)$ into feasible action space range. 

Because the weights of the convolutional reservoir computing layer are fixed, only the weight parameter $W^{\text{out}}$ requires training. We optimize $W^{\text{out}}$ by using CMA-ES, as in the world model. Therefore, it is possible to parallelize the training process and handle both discrete and continuous values as actions \cite{doi:10.1162/106365601750190398,DBLP:journals/corr/Hansen16a}. The process of optimizing $W^\text{out}$ by CMA-ES are followings:
\begin{enumerate}
  \item  Generate each solution candidate $W_{i}^{\text{out}}$ $(i=1,...,n)$ from a multivariate normal distribution $\mathcal{N}(m, \sigma^2C)$.
  \item  Create $n$ environments and workers. Each worker $\text{worker}_i$ $(i=1,...,n)$ implements the RCRC model and $W_i^{\text{out}}$ is set to the controller layer.
  \item  In each environment, each $\text{worker}_i$ plays $m$ episodes and in each episode, $\text{worker}_i$ receives a score $G_{i, j}$ $(j =1, ..., m)$ .
  \item  Update evolution paths with the score of each $W_i^{\text{out}}$ which is calculated by $G_i = 1/m\sum_{j=1}^m{G_{i, j}}$.
  \item  Update $m$, $\sigma$, $C$ by using evolution paths.
  \item  Repeat 1 to 5 until the convergence condition is satisfied or the specified number of repetitions are completed.
\end{enumerate}
In this process, $n$ means the number of solution candidates $W^{\text{out}}$ generated at each step. Each worker extracts features, takes the action in each independent environment, and obtains scores.  

\begin{figure}[t]
    \centering
    \subfigure[{\tt CarRacing-v0}
    ]{
      \centering
      \includegraphics[width=0.45\columnwidth]{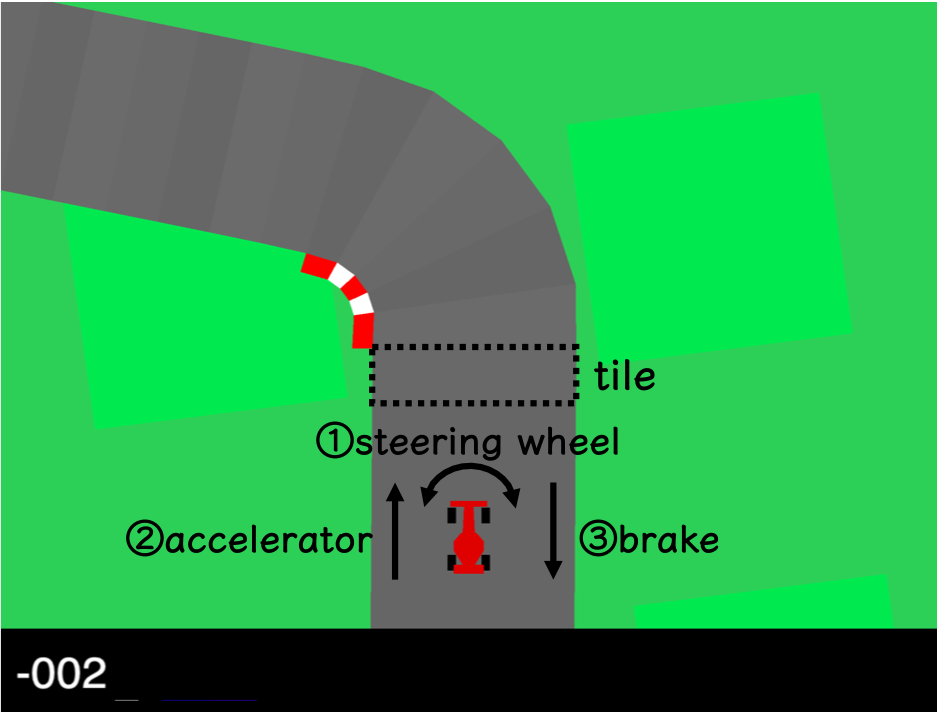}
      \label{fig:fig3a}
    }
    \subfigure[{\tt DoomTakeCover-v0}
    ]{
      \centering 
      \includegraphics[width=0.45\columnwidth]{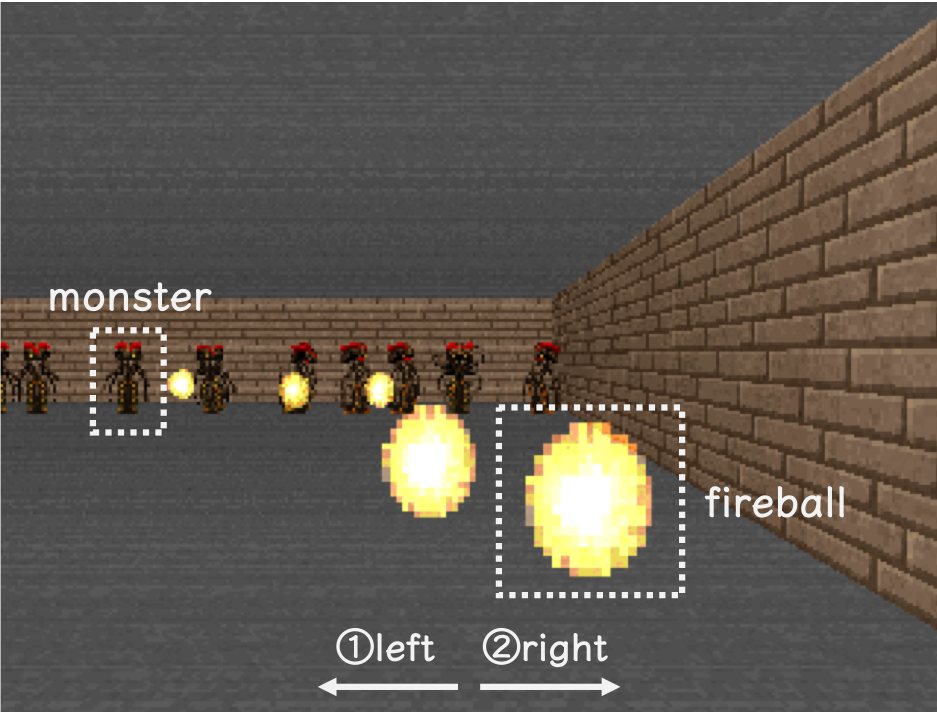}
      \label{fig:fig3b}
    }
    \caption{Example environment state images and actions in each environment.}
    \label{fig:fig3}
\end{figure}

\section{Experiments}
\subsection{Experiments Environments}
We evaluate the RCRC model in two famous RL tasks: {\tt CarRacing-v0} \cite{carracingv0} in OpenAI Gym \cite{1606.01540} and {\tt DoomTakeCover-v0} \cite{Kempka2016ViZDoom,DoomTakeCover} in ViZDoom \cite{wydmuch2018vizdoom}. {\tt CarRacing-v0} is a continuous action task and {\tt DoomTakeCover-v0} is a discrete action task. In both environments, we use the identical structure and weights convolutional reservoir computing layer as a feature extractor to evaluate the generalization ability and train only a single weight in the controller layer.

\subsubsection*{CarRacing-v0}
{\tt CarRacing-v0} \cite{carracingv0} is a car racing game environment that is known as a difficult continuous action task \cite{NIPS2018_7512}. The goal of this game is to go around the course without getting out by operating a car with three continuous actions: steering wheel, accelerator, and brake. The course is filled with tiles as shown in Figure~\ref{fig:fig3}\subref{fig:fig3a}. Each time the car passes a tile on the course, $1000 / N$ is added to the score. The scalar $N$ is the total number of tiles on the course. The course is randomly generated every time, and the total number of tiles in the course varies around 300. If all the tiles are passed, the total reward will be 1000, but it is subtracted by 0.1 for each frame. The episode ends when all the tiles are passed or when 1000 frames are played. If the player can pass all the tiles without getting out of the course, the reward will be above 900. The definition of ``solve" in this game is to get an average of 900 over 100 consecutive trials. 

\subsubsection*{DoomTakeCover-v0}
{\tt DoomTakeCover-v0} \cite{Kempka2016ViZDoom,DoomTakeCover} is a first-person perspective and 3D vision game. The goal of this game is to survive a long time with avoiding fireballs by operating a player with two discrete actions: move left and move right as shown in Figure~\ref{fig:fig3}\subref{fig:fig3b}. The fireballs are launched towards the player from the monsters and the survival time steps $\times$ 1 will be the score. Player has hit points and can withstand 1 or 2 times of fireballs hit, but if hit points has gone, it will be game over. The screen moves along with its own action, and the fireballs are launched from the backward of the screen toward the player, so it is necessary to recognize the depth of the screen to avoid the fireball. The episode ends when 2100 frames are passed. The definition of ``solve" in this game is to get an average of 750 over 100 consecutive trials.

\subsection{Procedure}
As previously stated, taking advantage of the RCRC model characteristic that feature extractor's weights are task-independent, we use an identical convolutional reservoir computing layer as a feature extractor in both tasks. 

\begin{figure}[t]
  \centering
  \includegraphics[keepaspectratio, width=0.8\columnwidth]{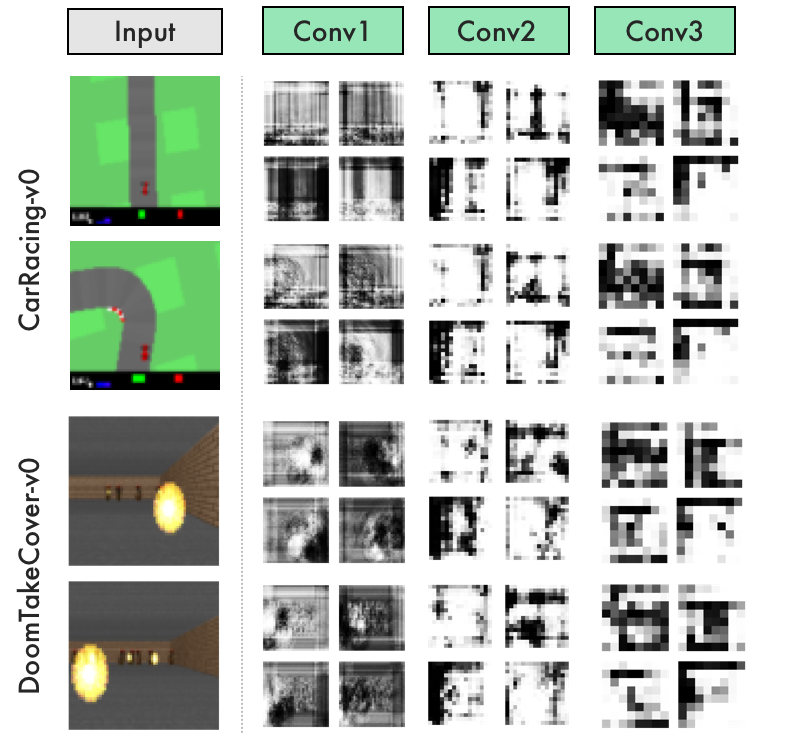}
  \caption{Examples of the visual features in each convolution layer. The features in the same column are extracted by the same network.}
  \label{fig:fig4}  
\end{figure}

We first resize the environment state image into 64 $\times$ 64 with 3 channels pixels and divided by 255 to restrict the each pixel value into $[0,1]$ as input to the convolutional reservoir computing layer. In the convolutional reservoir computing layer, we set 3 convolution layers and 1 dense layer. The filter sizes in the convolutional layers are 31, 14, and 6, and we set the number of the filters to 32, 64, and 128. All strides are set to 2. We set $D_\text{conv}$ and $D_\text{esn}$ to 512. The weights in convolution layers are sampled from Gaussian distributions $\mathcal{N}(0, 0.06^2)$; the both weights in the reservoir computing layer $W^\text{in}$ and $W$ are sampled from Gaussian distributions $\mathcal{N}(0, 0.1^2)$. In the reservoir computing layer, we set the leakage rate $\alpha$ to 0.8, the sparsity of $W$ to 0.8, and the spectral radius of $W$ to 0.95. All activation functions are set to $tanh$ which is often used in the ESN manner. The task-dependent parameters are only $W^\text{out}$ in the controller layer.  The size of $W^\text{out}$ which is the parameter be trained in a task-dependent manner is 3075 in {\tt CarRacing-v0} and 1025 in {\tt DoomTakeCover-v0}. The examples of the visual features extracted in each convolution layer are shown in Figure~\ref{fig:fig4}.

To get action $A_\text{car}(t)$ of {\tt CarRacing-v0}, as in the world model \cite{NIPS2018_7512}, we adjust $\tilde{A}_\text{car}(t)$ which is calculated by dot product of the extracted features and the weight in the controller layer, by the function $g_\text{car}$ as follows: 
\begin{eqnarray}
A_\text{car}(t) = g_\text{car}(\tilde{A}_\text{car}(t)) = 
\begin{cases}
    tanh\left(\tilde{A}_\text{car}^{\text{(1)}}(t)\right) \\
    [tanh\left(\tilde{A}_\text{car}^{\text{(2)}}(t)\right) + 1.0] / 2.0 \\
    clip[tanh\left(\tilde{A}_\text{car}^{\text{(3)}}(t)\right), 0, 1]
\end{cases}
\end{eqnarray}
where $\tilde{A}_\text{car}^{(i)}$ is $i$-th value in $\tilde{A}$ and $clip[x, \lambda_{\text{min}}, \lambda_{\text{max}}]$ is the function that limits the value of $x$ in range from $\lambda_{\text{min}}$ to $\lambda_{\text{max}}$ by clipping. Let $A_\text{car}^{(i)}$ be $i$-th value in $A$, the values $A_\text{car}^{(1)} \in [-1, 1], A_\text{car}^{(2)} \in [0, 1]$ and $A_\text{car}^{(3)} \in [0, 1]$ are correspond to steering wheel, brake and accelerator, respectively. 

To get action $A_\text{doom}(t)$ of {\tt DoomTakeCover-v0}, we adjust $\tilde{A}_\text{doom}(t)$ which is calculated by dot product of the extracted features and weight in the controller layer, by the function $g_\text{doom}$ as follows:
\begin{eqnarray}
A_\text{doom}(t) = g_\text{doom}(\tilde{A}_\text{doom}(t)) = 
\begin{cases}
    \text{left}  & (\tilde{A}_\text{doom}^{\text{(1)}}(t)\leq 0) \\
    \text{right}  &  (\tilde{A}_\text{doom}^{\text{(1)}}(t) > 0) 
\end{cases}
.
\end{eqnarray}

In the experiments, we use CMA-ES to optimize $W^{\text{out}}$ until 500-th generations, and set 16 workers $(n=16)$ for {\tt CarRacing-v0} and 32 workers $(n=32)$ for {\tt DoomTakeCover-v0}. Each worker is set to simulate over 8 randomly generated trials $(m=8)$, and updates $W^{\text{out}}$ with an average of these scores. In optimizing $W^{\text{out}}$ in {\tt DoomTakeCover-v0}, we didn't set the max simulation step to evaluate the actual playing ability. As in the world model \cite{NIPS2018_7512}, we evaluate the generalization ability of the models by the average score over 100 randomly created trials.  In generalization ability evaluation, we set the weight of the best worker which reached the best average score over 8 trials to the controller layer's weight.

To investigate the ability of network structures, we evaluate three models: the full RCRC model, the RCRC model that removes the reservoir computing layer (visual model), the RCRC model that has only one dense layer as feature extractor (dense model). The dense model uses flatten vector of 64 $\times$ 64 with 3 channels pixels as input and extracts visual features with no convolution. We set the weights of all models to random and fixed. The inputs to the controller layer of the visual model and the dense model are the $D_{\text{conv}}$-dimensional outputs from the dense layer shown in Figure~\ref{fig:fig2}.

\begin{figure*}[t]
    \centering
    \subfigure[{\tt CarRacing-v0}
    ]{
      \centering
      \includegraphics[width=0.47\textwidth]{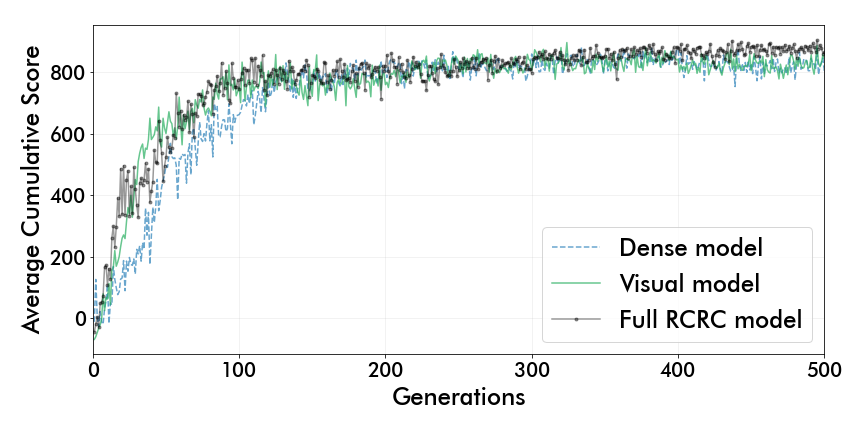}
      \label{fig:fig5a}
    }
    \subfigure[{\tt DoomTakeCover-v0}
    ]{
      \centering
      \includegraphics[width=0.47\textwidth]{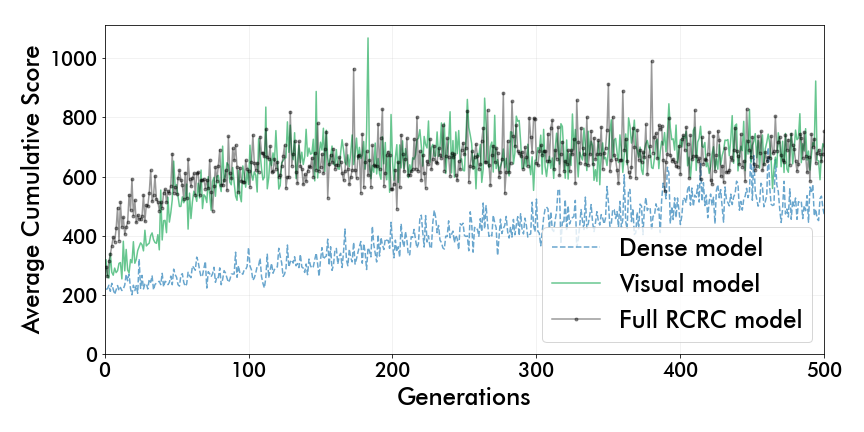}
      \label{fig:fig5b}
    }
    \caption{(a) The best average score over 8 randomly created trials among 16 workers in {\tt CarRacing-v0}. (b) The best average score over 8 randomly created trials among 32 workers in {\tt DoomTakeCover-v0}.}
    \label{fig:fig5}
\end{figure*}

\subsection{Results}
\subsubsection*{CarRacing-v0}
The best scores among 16 workers are shown in Figure~\ref{fig:fig5}\subref{fig:fig5a}. Each worker's score is evaluated as an average score over 8 randomly generated tracks.
Incredibly, the dense model reached an average score above 880 over 8 randomly generated tracks, and the visual model reached above 890. The dense model's score transition has higher volatility than the visual model's score transition. Furthermore, the visual model's score is less stable than the full RCRC model's score. These results shows that only one dense layer can extract visual features despite the fact that the weights are random and fixed, and the features extracted by the convolutional layers and the ESN improved scores. 

The generalization ability of the visual model and the full RCRC model which evaluated as an average score over 100 random trials are shown in Table~\ref{tab:table1}. The visual model which uses 512-dimensional visual features achieved 864 $\pm$ 79 which is better than the V model that uses 32-dimensional features extracted by VAE as input to controller layer in the world model. In addition, the full RCRC model reached 902 $\pm$ 21 which is comparable to state of the art approaches such as the world model approach \cite{NIPS2018_7512} and GA approach \cite{DBLP:journals/corr/abs-1906-08857}. Therefore the full RCRC model can be regarded as having ability to solve {\tt CarRacing-v0}. These results show the time-series features extracted by the ESN improves driving skill. 

\subsubsection*{DoomTakeCover-v0}
The best scores among 32 workers are shown in Figure~\ref{fig:fig5}\subref{fig:fig5b}. Each worker's score is evaluated as an average score over 8 randomly generated trials. While the dense model only improved score little by little, the visual model and the full RCRC model improved scores fast and reached above 750 in early steps. Therefore, it seems that it is difficult to express complex visual features such as the depth of screens with the dense layer alone, and the convolutional layer is effective. Although the full RCRC model reached a higher score than the visual model in early steps, the visual model achieved above 1000, which is higher than the full RCRC model. 

The generalization ability of the visual model and the full RCRC model which is evaluated as an average score over 100 random trials are shown in Table~\ref{tab:table2}. For comparison, the scores of the OpenAI Gym leaderboard \cite{DoomTakeCover} and the self-playing world model with different temperature parameter $\tau$ are listed. The temperature $\tau$ controls the variance of the next environment state prediction. A large $\tau$ means that the model predicts next environment state with high variance. On the other hand, if $\tau$ sets to 0, the model predicts the next environment state deterministically.

The full RCRC model reached 922 $\pm$ 450 and the visual model achieved 832 $\pm$ 483. They couldn't reach the best score of the self-playing world model 1092 $\pm$ 556, but they greatly exceed above 750 which means ``solved" the task. Although the best score of the visual model exceeds that of the full RCRC model in the parameter optimization process, at average score over 100 random trials, the visual model's score is lower than the full RCRC model's one. These results suggests that using time-series features extracted by the RCRC model improves the generalization ability. 

Furthermore, the RCRC model's feature extractor which is completely identical between tasks has generalization ability to solve both tasks by training only a linear transformation from the extracted features to the actions, despite the fact that the network's weights are set to random and fixed. 
\begin{table}[t]
 \centering
 \caption{{\tt CarRacing-v0} scores of various methods.\\}
 \vspace*{0.1cm}
 \begin{tabular}{c}
    \centering
    \small
      \begin{tabular}{lll}
      \hline 
      Method     & AVG. Score  \\
      \hline
      DQN \cite{DQNscore} & 343  $\pm$ 18     \\  
      A3C \cite{A3Ccont} & 591 $\pm$ 45  \\
      \hspace{-1em}
      \begin{tabular}{l}
      World model with random MDN-RNN \\ \qquad\cite{Corentin2018rnnfix}           
      \end{tabular}
      &   870  $\pm$ 120 \\ 
      
      World model \cite{NIPS2018_7512} \\
       \qquad V model     &    632 $\pm$ 251 \\
       \qquad World model     &   {\bf 906}  $\pm$ 21 \\
      GA \cite{DBLP:journals/corr/abs-1906-08857}     &   {\bf 903}  $\pm$  73 \\    
      \hline
      RCRC model (Visual model)  &    864  $\pm$  79\\     
      RCRC model     &  {\bf 902}  $\pm$  21 \\
      \hline 
    \end{tabular}
    \end{tabular}
  \label{tab:table1}
\end{table}

\begin{table}[t]
 \centering
 \caption{{\tt DoomTakeCover-v0} scores of various methods.}
 \vspace*{0.1cm}
 \begin{tabular}{c}
    \centering
    \small
    \vspace*{0.1cm} 
    \begin{tabular}{lll}
      \hline 
      Method     & AVG. Score  \\
      \hline   
      Gym Leader \cite{DoomTakeCover}  &  {\bf 820}  $\pm$ 58 \\
      World model \cite{NIPS2018_7512} \\
       \qquad $\tau=0.10$     &   193  $\pm$ 58 \\
       \qquad $\tau=0.50$     &   196  $\pm$ 50 \\
       \qquad $\tau=1.00$     &   {\bf 868} $\pm$ 511 \\
       \qquad $\tau=1.15$     &   {\bf 1092}  $\pm$ 556 \\
       \qquad $\tau=1.30$     &   {\bf 753}  $\pm$ 139 \\
      \hline
      RCRC model (Visual model)  &    {\bf 832}  $\pm$  483\\     
      RCRC model     &  {\bf 922}  $\pm$  450 \\
      \hline 
    \end{tabular}
    \end{tabular}
  \label{tab:table2}
\end{table}

\section{Conclusions and Discussions}

In this study, we focused on extracting features that sufficiently express the environment state, rather than those that are trained to get higher rewards. To this end, we developed a novel approach called RCRC model which using fixed random-weight CNN and a novel ESN-based method, respectively, extracts visual features from environment state images and time-series features from transitions of visual features. This model architecture results in highly practical features that omit the training process of the feature extractor and reduce computational costs, and there is no need to store large volumes of data.  Surprisingly, extracted features are expressive enough to solve multiple RL tasks with training only a linear transformation of those features, despite the fact that it used the completely identical feature extractor. These results bring us to the conclusion that network structures themselves, such as CNN and ESN, have the capacity to extract features, and the RCRC model has generalization ability to express various environments and solve RL tasks. 

Although the RCRC model is not suitable for the tasks that are hard to simulate because it optimizes parameters by the simulated score with current parameters, it has the potential to make RL widely available. Recently, many RL models have achieved high performance in various tasks, but most of them have high computational costs and often require significant time for training. This makes the introduction of RL inaccessible to many. However, by using the RCRC model anyone can build high-performance models fast with much lower computational costs. In addition, the RCRC model can handle a wide range of actions, and even when the environment changes, training can be performed without any pre-training. Therefore, the RCRC model can be used easily by anyone to apply to various environments. 

While in {\tt CarRacing-v0}, the full RCRC model reached a comparable score to the best score of the world model, the full RCRC model couldn't reach in {\tt DoomTakeCover-v0}. The world model uses VAE and MDN-RNN, and can extract probabilistic features based on the assumption of multiple future environment states, but the RCRC model can only extract deterministic features by actual image input. In {\tt DoomTakeCover-v0}, the environment state images are first-person view, and not all states can be observed. Therefore, it seems that the world model can get a higher score.

As a further improvement, there is a possibility that the score can be improved by ensembling multiple features that are extracted by multiple convolutional reservoir computing layers as in the ESN \cite{massar2013mean}. The convolutional reservoir computing layer uses random weights generated from Gaussian distributions. Therefore, it can easy to obtain multiple independent features by using different random seeds. In fact, we checked that multiple convolutional reservoir computing layers based on different random seeds can solve both tasks with training a single weight in the controller layer. Thus, it seems effective ensembling such features. In addition, assigning more workers and using state of the art CNNs and RNNs with fixed random-weight extractor have the possibility to improve performance.

In future work, we consider making predictions from previous extracted features and actions to the next ones to be an important and promising task. Because the ESN was initially proposed to predict complex time-series, it can be assumed to have capacity to predict next features. If this task is achieved, it can self-simulate RL tasks by making iterative predictions from an initial state. This will help to broaden the scope of RL applications.
\section{Acknowledgements}
The authors are grateful to Takuya Yaguchi for the discussions on reinforcement learning. We also thank Hiroyasu Ando for helping us to improve the manuscript.
\fontsize{9.8pt}{10.8pt} \selectfont
\bibliographystyle{aaai}
\bibliography{references}

\end{document}